\documentclass{article}
\usepackage{spconf,amsmath,epsfig}
\usepackage{amsmath,amssymb}
\usepackage[table]{xcolor} % 用于颜色和表格颜色
\usepackage{pifont}
\usepackage{multirow, multicol}
\usepackage{booktabs}
\usepackage{mathrsfs}

\title{Toward distortion-aware change detection in realistic scenarios}
\name{Yitao Zhao, Heng-Chao Li\sthanks{This work is supported by National Natural Science Foundation of China under Grants 62271418 and 61871335.
}, Nanqing Liu, Rui Wang}
\address{School of Information Science and Technology, Southwest Jiaotong University, \\ Chengdu, China}
\begin{document}
%\ninept
%
\maketitle
\begin{abstract}
In the conventional change detection (CD) pipeline, two manually registered and labeled remote sensing datasets serve as the input of the model for training and prediction. However, in realistic scenarios, data from different periods or sensors could fail to be aligned as a result of various coordinate systems. Geometric distortion caused by coordinate shifting remains a thorny issue for CD algorithms. In this paper, we propose a reusable self-supervised framework for bitemporal geometric distortion in CD tasks. The whole framework is composed of Pretext Representation Pre-training, Bitemporal Image Alignment, and Down-stream Decoder Fine-Tuning. With only single-stage pre-training, the key components of the framework can be reused for assistance in the bitemporal image alignment, while simultaneously enhancing the performance of the CD decoder. Experimental results in 2 large-scale realistic scenarios demonstrate that our proposed method can alleviate the bitemporal geometric distortion in CD tasks.
\end{abstract}


\section{Introduction}
\label{sec:intro}
Change detection (CD) is an essential approach for surface observation with remote sensing data, which aims to extract and highlight semantic features of changed objects from two or more remote sensing images with overlapping acquisition areas. Due to the ability of rapid detection of changing features in large-scale areas, CD methods have been widely applied in urban planning, farmland monitoring, and emergency and emergency response \cite{liu2019review, liu2021afdet}.

Conventional CD methods mainly intuitively acquire changing results by the direct comparison of pixels, image features, or pre-classification results \cite{johnson1998change, celik2009, kuncheva2014}. However, such methods are susceptible to the negative impact of the representation differences between bitemporal images \cite{zhao2022comparative, liu2023transformation}. With the emergence of massive remote sensing images, the change detection methods have gradually evolved from the initial algebra-based methods to the recent deep learning-based methods. Deep learning-based CD methods extract bitemporal features through the backbone network and perform feature fusion to obtain the prediction results. These methods significantly enhance the robustness of the fake change caused by the imaging representation difference \cite{daudt2018fully, fang2021snunet, chen2021remote, lei2023ultralightweight}.

Beyond the representation differences between bitemporal images, an external factor neglected by most researchers is the coordinate misalignment between the source images. In realistic scenarios, image registration is typically conducted based on the geographic metadata provided by the sensors or manually selected control points. The subsequent CD tasks are performed with the aligned bitemporal images. Nevertheless, an additional workload is introduced by this approach, and the entire workflow is fragmented. \textit{Can we combine the image alignment and change detection tasks to reduce redundancy in existing methods?} Motivated by this, we rethink the entire CD workflow. 

In this paper, we propose a change detection framework for realistic scenarios with geometric distortions. The whole framework is composed of Pretext Representation Pre-training, Bitemporal Image Alignment, and Down-stream Decoder Fine-Tuning. First, the unlabeled data is utilized for self-supervised learning to obtain the pretext representation pre-training. Then the alignment auxiliary head is frozen to promote the image alignment between bitemporal images. Finally, the reusable encoder is transferred for downstream CD decoder fine-tuning. 

% The rest of this paper is organized as follows. The details of the proposed framework is described in Section \ref{sec:method}. Section \ref{sec:experiment} explains the related experimental settings as well as the comparative results. Finally, the summary and perspectives are presented in Section \ref{sec:conclusion}.
\vspace{-0.2cm}
\section{Method}
\label{sec:method}

\subsection{Overview}
\begin{figure*}[!htbp]
     \vspace{-0.5cm}
	\centering
        \scalebox{0.9}{
	\includegraphics[width=\linewidth]{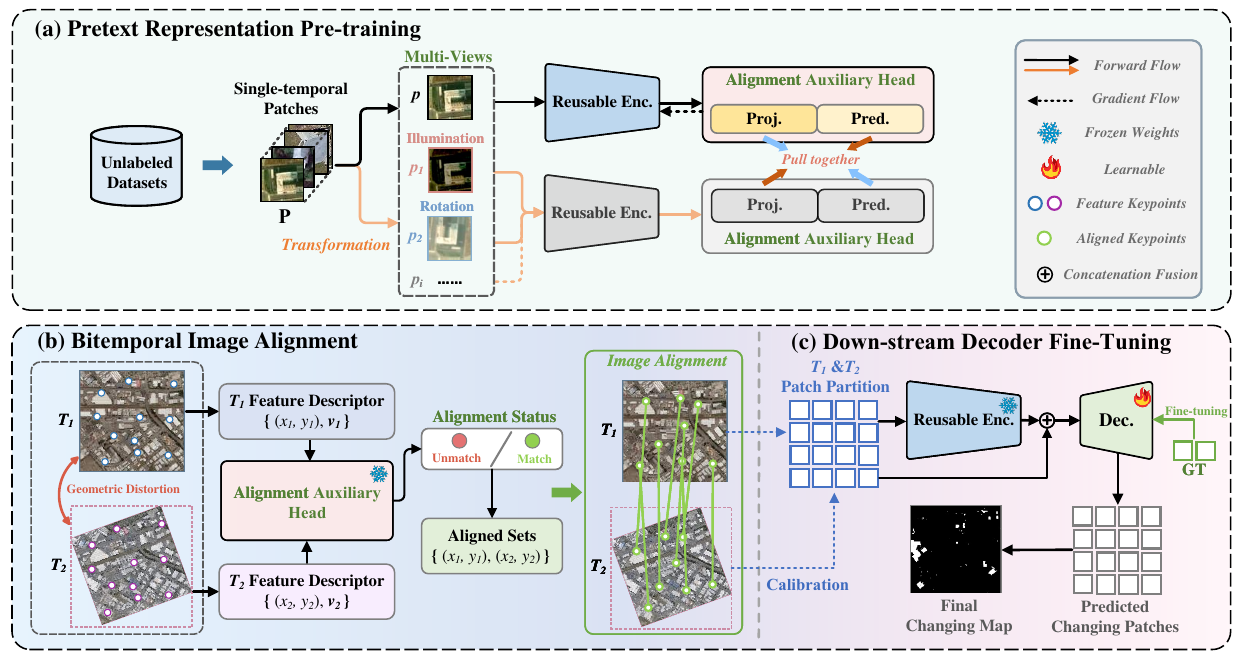}}
      \vspace{-0.2cm}
	\caption{Workflow of the proposed framework for RS CD tasks. (a) is the Pretext Representation Pre-training; (b) represents the Bitemporal Image Alignment; (c) is the Down-stream Decoder Fine-Tuning.}
	\label{fig:workflow}
     \vspace{-0.5cm}
\end{figure*}

The overall workflow of the proposed framework is illustrated in Figure \ref{fig:workflow}. This framework is structured into three distinct stages:  (a) Pretext Representation Pre-training; (b) Bitemporal Image Alignment; (c) Down-stream Decoder Fine-Tuning. Subsequent sections will provide an in-depth explanation of each component's role.

% ------ Workflow -------

\vspace{-0.2cm}
\subsection{Pretext Representation Pre-training}
The procedure of the Pretext Representation Pre-training is shown in Figure \ref{fig:workflow} (a). In this part, only unlabeled single-temporal patches are required for contrastive pre-training. Given a single-temporal patch set $\mathbf{P}$ sampled from the unlabeled dataset, a random sample in $\mathbf{P}$ is denoted as $p$. We perform augmentations to the sample $p$, including illumination transformation, random rotation, etc., to obtain the transferred sets with $M$ samples $ \left\{ p_{1}, p_{2}, \cdots, p_{i}, \cdots, p_{M} \right\}$. 

In this framework, the initial sample $p$ is input into the top branch, which consists of a reusable encoder and an alignment auxiliary head. Concurrently, the transferred samples are processed through the bottom branch, equipped with identical architecture. The cross-similarity between the outputs of the two alignment auxiliary heads is then computed. This step is crucial for enabling the model to learn deep semantic features from unlabeled data. Notably, during backward propagation, the gradient flow is truncated in the bottom branch. The training process is mathematically represented as follows:
\begin{equation}
\vspace{-0.15cm}
\mathop{\arg\min}{\theta} \sum\nolimits_{i=1}^{M} \mathcal{L} (f_{\theta}(p), f_{\theta}(p_{i}))
\vspace{-0.1cm}
\end{equation}
Here, $f_{\theta}$ denotes the combined functionality of the Reusable Encoder and Alignment Auxiliary Head, with $\theta$ representing their parameters. $p$ is the initial sample, and $p_{i}$ represents each corresponding transferred sample. The objective is to minimize the distance between the outputs of the twin branches, which is expected to endow the pre-trained encoder and alignment auxiliary head with robust generalization capabilities.
% minize (1/M)xigema_(i=1, M)(L(p, p_i))
\vspace{-0.2cm}
\subsection{Bitemporal Image Alignment}
In realistic scenarios, initial images acquired by various sensors may not be aligned, and the existing geometric distortion poses impediments to subsequent tasks such as change detection or time series analysis. As shown in Figure \ref{fig:workflow} (b). We employ the well-trained alignment auxiliary head to promote the registration between the bitemporal images. Suppose $\mathbf{D_1} = \left\{ (x_1, y_1), \mathit{v_1} \right\}$ represents a feature descriptor extracted from te $T_1$ image, and $\mathbf{D_2} = \left\{ (x_2, y_2), \mathit{v_2} \right\}$ is the one from $T_2$ image. $(x_i, y_i)$ represents the image coordinate, and $\mathit{v_1}$ is the corresponding feature vector. We believe that the frozen alignment auxiliary head possesses the ability to discriminate the similarity between the bitemporal feature descriptors as follows:

%  S = Sort(D(vi, vj))
\begin{small}
    \begin{equation}
    \left\{\begin{array}{l}
    \vspace{1.5ex}
    % \vspace{-0.15cm}
        \mathcal{D} = \mathbf{sort}[\mathit{d}(\mathit{v_{i}}, \mathit{v_{j}})], \quad (\mathit{v_{i}}\in\mathbf{D_1}, \mathit{v_{j}}\in\mathbf{D_2})
        \\
        \mathcal{S}_{match} = \left\{\mathcal{D}_{k} | \mathcal{D}_{k} \geq \tau \right\}
    \end{array}\right.
   \vspace{-0.15cm}
    \end{equation}
\end{small}
where $\mathcal{D}$ is the distance set sorted by the distance between $\mathit{v_{i}}$ and $\mathit{v_{j}}$, and $\mathit{d}$ is the cosine distance. $\tau$ is the threshold for filtering the distance set $\mathcal{D}$ to obtain the matching sets $\mathcal{S}_{match}$. Filtered by the alignment auxiliary head, the matching keypoint sets $\left\{ (x_{i}, y_{i}), (x_{j}, y_{j}) \right\}$ will be selected for the bitemporal image alignment by perspective calibration.

\begin{table*}[!htp]
    \centering    
    \caption{Comparative results of the selected CD baselines in the changing scenarios.}
    \label{tab:results}
    \renewcommand{\arraystretch}{1.3} % 调整行间距
    \setlength{\tabcolsep}{8pt} % 调整列间距
    \scriptsize{}
    \begin{tabular}{c|c|cccc|cccc}
        \toprule
        \multirow{2}{*}{\textbf{External Alignment}} & \multirow{2}{*}{\textbf{Method}} & \multicolumn{4}{c|}{\textbf{Changing Scenario 1}} & \multicolumn{4}{c}{\textbf{Changing Scenario 2}} \\
        \cmidrule{3-10}
        & & \textbf{Pre} & \textbf{Rec} & \textbf{F1} & \textbf{IoU} & \textbf{Pre} & \textbf{Rec} & \textbf{F1} & \textbf{IoU} \\
        \midrule
        \multirow{4}{*}{\textit{Require}} & FC-Siam-Conc \cite{daudt2018fully} & 14.25 & 82.01 & 24.27 & 13.81 & 52.31 & 83.91 & 64.45 & 47.54 \\
        % \cmidrule{2-10}
        & FC-Siam-Diff \cite{daudt2018fully} & 19.24 & 78.92 & 30.93 & 18.30 & 44.56 & 85.30 & 58.54 & 41.38 \\
        % \cmidrule{2-10}
        & SNUNet \cite{fang2021snunet} & 68.03 & 65.45 & 66.72 & 50.06 & 82.03 & 73.51 & 77.54 & 63.32 \\
        % \cmidrule{2-10}
        & USSFCNet \cite{lei2023ultralightweight} & 87.82 & 90.22 & 89.01 & 80.18 & 84.07 & 85.49 & 84.78 & 73.58 \\
        \midrule
        \rowcolor{green!4}
        \textit{No need} & \textbf{Ours} & \textbf{89.47} & \textbf{95.25} & \textbf{92.27} & \textbf{85.65} & \textbf{84.87} & \textbf{91.11} & \textbf{87.88} & \textbf{78.37} \\
        \bottomrule
    \end{tabular}
    \vspace{-0.5cm}
\end{table*}

\subsection{Down-stream Decoder Fine-tuning}
Once the bitemporal images are aligned to a common coordinate system, a significant reduction in geometric distortion can be achieved. This alignment allows for the partitioning of patches on the registered $T_1$ and calibrated $T_2$ images, facilitating the creation of inference patch sets. In the downstream Change Detection (CD) task, the pre-trained encoder is reused to augment the feature extraction phase, enabling more accurate semantic localization. At this stage, only a minimal amount of labeled data is needed to fine-tune the downstream CD decoder. Following the fine-tuning of the decoder, the inference patch sets are processed through the reusable encoder. The deep features extracted by the frozen encoder are then concatenated with the initial embeddings. This combined data serves as the input for the decoder, enhancing the overall feature representation for the CD task.

\begin{equation}
    \left\{\begin{array}{l}
        % \left\{ \right\}
        \mathit{f}_{T_{1}} = \mathit{Concat}[T_1, \mathbf{Enc}(T_1)]
        \\
        \mathit{f}_{T_{2}} = \mathit{Concat}[T_2, \mathbf{Enc}(T_2)]
    \end{array}\right.
\end{equation}

Finally, the changing map will be obtained based on the predicted changing patches. The calculation process can be illustrated as follows:

% minimize(Dec(Concat(T1, Enc(T1)), Concat(T2, Enc(T2))), GT)
\begin{equation}
    \vspace{-0.3cm}
    \mathbf{minimize} \quad \mathcal{L}(\mathbf{Dec}(\mathit{f}_{T_{1}}, \mathit{f}_{T_{2}}), GT)
\end{equation}

where \textbf{Dec} and \textbf{Enc} represent the down-stream decoder and reusable encoder. CM is the predicted changing map, and GT is the corresponding ground truth. $\mathit{f}_{T_{1}}$ and $\mathit{f}_{T_{2}}$ are the fused features.

\section{Experimental Results}
\label{sec:experiment}
\subsection{Related Settings}
\textbf{Datasets}: For the Pretext Representation Pre-training phase, we conducted the optimization process on the unlabeled WHU-CD dataset \cite{Ji_whucd}. The data patches were cropped to a resolution of $256 \times 256$ pixels. To evaluate our proposed framework, we utilized the source satellite images from \cite{Ji_whucd}. Specifically, we selected two changing regions measuring $5649 \times 5433$ and $4372 \times 5383$ pixels. We then applied random perspective transformations to one image in each region to simulate realistic change scenarios.

\noindent \textbf{Evaluation Metrics}: The performance of the proposed framework was assessed using four metrics: Precision (Pre), Recall (Rec), F1-Score, and Intersection over Union (IoU).

\noindent \textbf{Implementation Details}: The Pretext Representation Pre-training was executed for 100 epochs with a batch size of 16, using the SGD optimizer with a momentum of 0.9 and weight decay of 0.0001. For the downstream decoder fine-tuning, we set the batch size to 8. The parameters were optimized using the AdamW optimizer with a momentum of 0.999 and a weight decay of 0.01.

% \vspace{-0.2cm}
\subsection{Quantitative Analysis}
Based on the selected evaluation metrics, we perform prediction in the entire changing scenario. To the best of our knowledge, few existing CD methods cover the impact of geometric distortion. However, for a fair comparison, we apply an additional scenario registration step to the comparative CD methods. In consideration of the efficiency of the registration procedure, we choose the SIFT algorithm \cite{lowe2004distinctive}, which is widely utilized in realistic scenarios due to its stable performance, to collaborate with the CD methods.

The comparative results in the selected changing scenario are listed in Table \ref{tab:results}. Through the vertical comparison, we can observe that the comparative methods present a significant performance degradation in the selected changing scenarios. We argue that since most of the regions in the experimental scenario remain unchanged, only a few number of buildings undergo major changes. Therefore, during the optimization of the baseline methods, the large quantity of unchanged samples leads to a deviation of the parameters from the optimal solution. In addition, the proposed method does not require external image alignment steps, realizing the alignment and change detection to be integrated. In contrast, other baselines require manual or specially designed alignment before training, thereby inflicting additional workload.

% ------ Visualization -------
\begin{figure*}[!htp]
	\centering
     \vspace{-0.5cm}
        \scalebox{0.9}{
	\includegraphics[width=\linewidth]{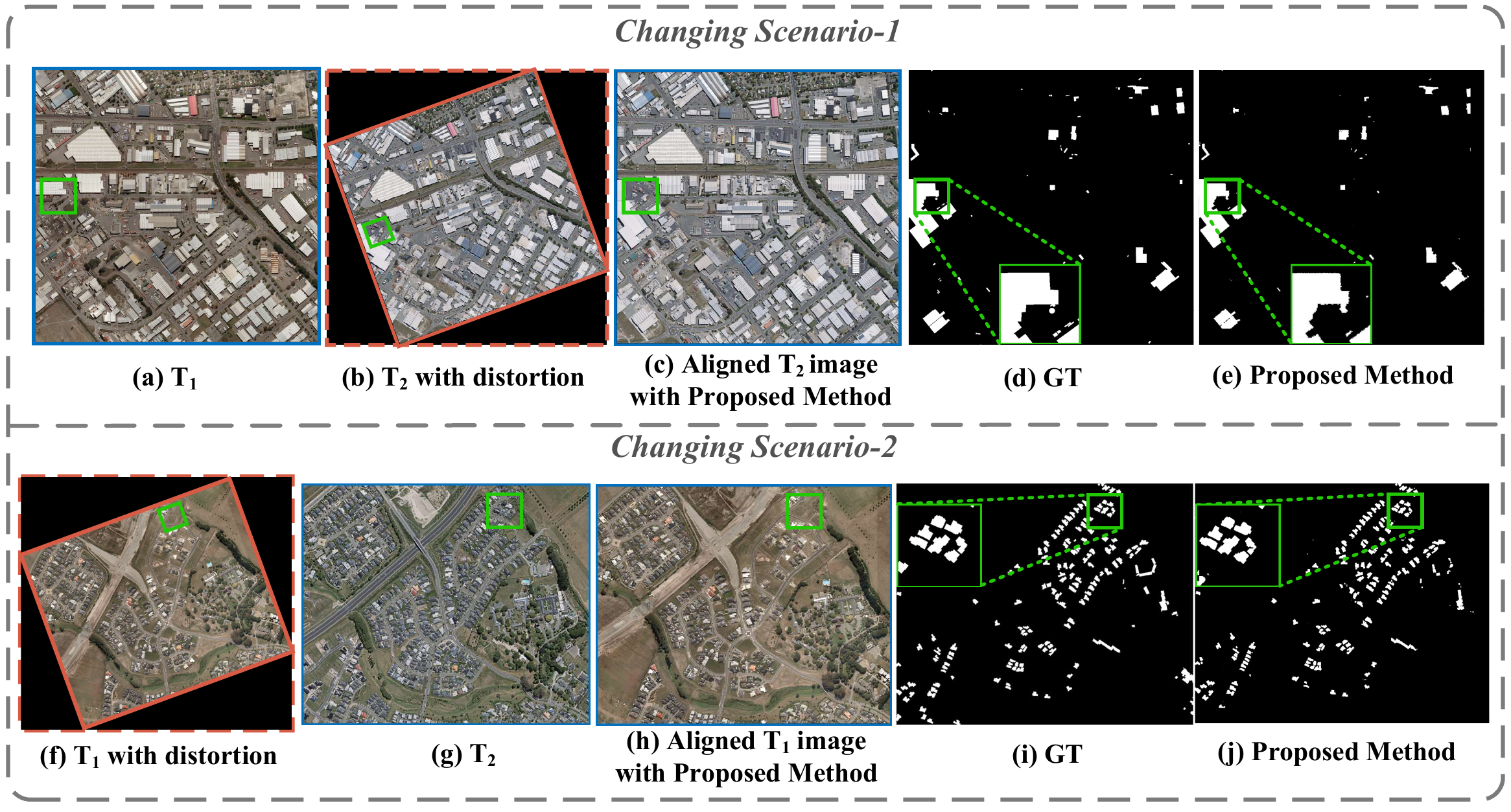}}
      \vspace{-0.4cm}
	\caption{Visualization of the proposed framework for the experimental scenarios with significant geometric distortion. }

	\label{fig:visual}
     \vspace{-0.4cm}
\end{figure*}

\vspace{-0.2cm}
\subsection{Visualization Analysis}
% We perform visualization for the experimental regions in order to illustrate the effectiveness of the proposed method. 
The visualization results in presented in Figure \ref{fig:visual}. Significant geometric distortion can be observed between the original $T_1$ images and $T_2$ images. The aligned result of the $T_2$ image is illustrated in Figure \ref{fig:visual}-(c) and (h). With the calibration of the Alignment auxiliary head, the distortion between the bitemporal images could be largely alleviated, which provides a solid basis for subsequent CD tasks. The change detection result of the proposed method is displayed in Figure \ref{fig:visual}-(e) and (i). Through the holistic comparison, we find that the proposed method excels in extracting changed regions, even in the presence of category imbalance. It demonstrates that the pre-training procedure of pretext representation provides rich deep semantic features for the down-stream decoder fine-tuning. Through the zoom-in view presented in Figure \ref{fig:visual}-(e) and (i) we notice that the proposed method is capable of accurately capturing the details of the changing objects.

\section{Conclusion}
\label{sec:conclusion}
In this paper, we propose a reusable self-supervised framework to handle the bitemporal geometric distortion in remote sensing CD tasks. The proposed framework is organized in a simple structure, containing the Pretext Representation Pre-training, Bitemporal Image Alignment, and Down-stream Decoder Fine-tuning. Specifically, the Alignment Auxiliary Head is assigned to deal with the geometric distortion between bitemporal images. The pre-trained encoder is reused to enhance the performance of feature extraction. The experimental results in the realistic scenario demonstrate the effectiveness of our proposed method.

% \vfill
% \pagebreak

\vspace{-0.2cm}
\bibliographystyle{IEEEbib}
\small
% \footnotesize
\bibliography{ref/ref}

\begin{thebibliography}{10}

\bibitem{liu2019review}
Sicong Liu, Daniele Marinelli, Lorenzo Bruzzone, and Francesca Bovolo,
\newblock ``A review of change detection in multitemporal hyperspectral images: Current techniques, applications, and challenges,''
\newblock {\em IEEE Geoscience and Remote Sensing Magazine}, vol. 7, no. 2, pp. 140--158, 2019.

\bibitem{liu2021afdet}
Nanqing Liu, Turgay Celik, Tingyu Zhao, Chao Zhang, and Heng-Chao Li,
\newblock ``Afdet: {Toward More Accurate} and{ Faster Object Detection} in {Remote Sensing Images},''
\newblock {\em IEEE Journal of Selected Topics in Applied Earth Observations and Remote Sensing}, vol. 14, pp. 12557--12568, 2021.

\bibitem{johnson1998change}
R.D. Johnson and E.S. Kasischke,
\newblock ``Change vector analysis: A technique for the multispectral monitoring of land cover and condition,''
\newblock {\em International Journal of Remote Sensing}, vol. 19, no. 3, pp. 411--426, 1998.

\bibitem{celik2009}
Turgay Celik,
\newblock ``Unsupervised {Change Detection} in {Satellite Images Using Principal Component Analysis} and {$k$-Means Clustering},''
\newblock {\em IEEE Geoscience and Remote Sensing Letters}, vol. 6, no. 4, pp. 772--776, 2009.

\bibitem{kuncheva2014}
Ludmila~I. Kuncheva and William~J. Faithfull,
\newblock ``{PCA Feature Extraction} for {Change Detection} in {Multidimensional Unlabeled} data,''
\newblock {\em IEEE Transactions on Neural Networks and Learning Systems}, vol. 25, no. 1, pp. 69--80, 2014.

\bibitem{zhao2022comparative}
Yitao Zhao, Turgay Celik, Nanqing Liu, and Heng-Chao Li,
\newblock ``A {Comparative Analysis} of {GAN-Based Methods} for {SAR-to-Optical Image Translation},''
\newblock {\em IEEE Geoscience and Remote Sensing Letters}, vol. 19, pp. 1--5, 2022.

\bibitem{liu2023transformation}
Nanqing Liu, Xun Xu, Turgay Celik, Zongxin Gan, and Heng-Chao Li,
\newblock ``Transformation-invariant {Network} for {Few-Shot Object Detection} in {Remote-Sensing Images},''
\newblock {\em IEEE Transactions on Geoscience and Remote Sensing}, vol. 61, pp. 1--14, 2023.

\bibitem{daudt2018fully}
Rodrigo~Caye Daudt, Bertr Le~Saux, and Alexandre Boulch,
\newblock ``Fully convolutional siamese networks for change detection,''
\newblock in {\em 2018 25th IEEE International Conference on Image Processing (ICIP)}. IEEE, 2018, pp. 4063--4067.

\bibitem{fang2021snunet}
Sheng Fang, Kaiyu Li, Jinyuan Shao, and Zhe Li,
\newblock ``{SNUNet-CD}: A densely connected siamese network for change detection of {VHR} images,''
\newblock {\em IEEE Geoscience and Remote Sensing Letters}, vol. 19, pp. 1--5, 2021.

\bibitem{chen2021remote}
Hao Chen, Zipeng Qi, and Zhenwei Shi,
\newblock ``Remote sensing image change detection with transformers,''
\newblock {\em IEEE Transactions on Geoscience and Remote Sensing}, vol. 60, pp. 1--14, 2021.

\bibitem{lei2023ultralightweight}
Tao Lei, Xinzhe Geng, Hailong Ning, Zhiyong Lv, Maoguo Gong, Yaochu Jin, and Asoke~K Nandi,
\newblock ``Ultralightweight {Spatial}--{Spectral Feature Cooperation Network} for {Change Detection} in {Remote Sensing Images},''
\newblock {\em IEEE Transactions on Geoscience and Remote Sensing}, vol. 61, pp. 1--14, 2023.

\bibitem{Ji_whucd}
Shunping Ji, Shiqing Wei, and Meng Lu,
\newblock ``Fully {Convolutional} {Networks} for {Multisource Building Extraction From} an {Open Aerial} and {Satellite Imagery Data Set},''
\newblock {\em IEEE Transactions on Geoscience and Remote Sensing}, vol. 57, no. 1, pp. 574--586, 2019.

\bibitem{lowe2004distinctive}
David~G Lowe,
\newblock ``Distinctive image features from scale-invariant keypoints,''
\newblock {\em International journal of computer vision}, vol. 60, pp. 91--110, 2004.

\end{thebibliography}
\end{document}